\documentclass[11pt]{article}

\usepackage[final]{acl}

\usepackage{times}
\usepackage{latexsym}

\usepackage[T1]{fontenc}
\usepackage[utf8]{inputenc}

\usepackage{microtype}

\usepackage{inconsolata}

\usepackage{graphicx}
\usepackage{amssymb}
\usepackage{enumitem}
\usepackage{amsmath}
\usepackage{booktabs} 
\usepackage{multirow}
\usepackage{fontawesome}
\usepackage{booktabs}
\usepackage{multirow}
\usepackage{colortbl}
\usepackage{xcolor}
\usepackage{tabularx}
\usepackage{adjustbox}
\usepackage{makecell} % 用于表头换行

\title{\raisebox{-1.5mm}{\includegraphics[width=0.05\textwidth]{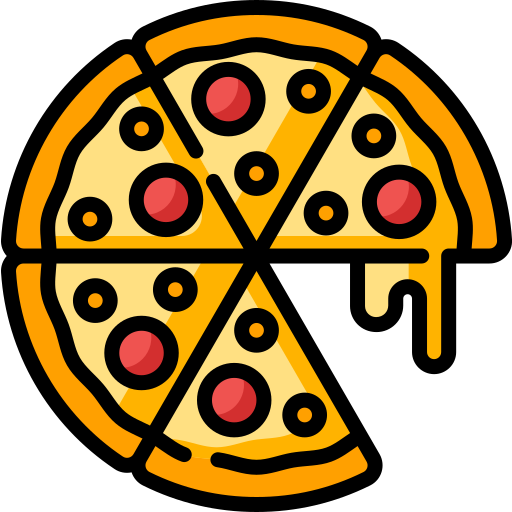}} DiningBench: A Hierarchical Multi-view Benchmark for Perception and Reasoning in the Dietary Domain}

\author{First Author \\
  Affiliation / Address line 1 \\
  Affiliation / Address line 2 \\
  Affiliation / Address line 3 \\
  \texttt{email@domain} \\\And
  Second Author \\
  Affiliation / Address line 1 \\
  Affiliation / Address line 2 \\
  Affiliation / Address line 3 \\
  \texttt{email@domain} \\}

\author{
   Song Jin\textsuperscript{1,2}\thanks{\ \  Equal contribution to this work.}, Juntian Zhang\textsuperscript{1}\footnotemark[1],
   \textbf{Xun Zhang}\textsuperscript{2}\thanks{\ \ Project leader.}, \\ \textbf{Zeying Tian}\textsuperscript{2}, \textbf{Fei Jiang}\textsuperscript{2}, \textbf{Guojun Yin}\textsuperscript{2}, \textbf{Wei Lin}\textsuperscript{2}, \textbf{Yong Liu}\textsuperscript{1}\thanks{\ \ Corresponding authors.},
  \textbf{Rui Yan}\textsuperscript{3}\footnotemark[3] \\
  \textsuperscript{1}Gaoling School of Artificial Intelligence, Renmin University of China, \\ \textsuperscript{2}Meituan, \textsuperscript{3}Wuhan University\\
  jinsong8@ruc.edu.cn
}

\begin{document}
\maketitle
\begin{abstract}

Recent advancements in Vision-Language Models (VLMs) have revolutionized general visual understanding. However, their application in the food domain remains constrained by benchmarks that rely on coarse-grained categories, single-view imagery, and inaccurate metadata. To bridge this gap, we introduce DiningBench, a hierarchical, multi-view benchmark designed to evaluate VLMs across three levels of cognitive complexity: \textit{Fine-Grained Classification}, \textit{Nutrition Estimation}, and \textit{Visual Question Answering}. Unlike previous datasets, DiningBench comprises 3,021 distinct dishes with an average of 5.27 images per entry, incorporating fine-grained ``hard'' negatives from identical menus and rigorous, verification-based nutritional data. We conduct an extensive evaluation of 29 state-of-the-art open-source and proprietary models. Our experiments reveal that while current VLMs excel at general reasoning, they struggle significantly with fine-grained visual discrimination and precise nutritional reasoning. Furthermore, we systematically investigate the impact of multi-view inputs and Chain-of-Thought reasoning, identifying five primary failure modes. DiningBench serves as a challenging testbed to drive the next generation of food-centric VLM research: \faGithub~\href{https://github.com/meituan/DiningBench}{DiningBench}.

\end{abstract}

\section{Introduction}

Food is fundamental to human existence, extending beyond mere sustenance to encompass culture, health, and lifestyle. With the rapid advancement of Vision-Language Models (VLMs), the potential for AI-assisted food analysis has grown exponentially, from automated dietary logging to intelligent kitchen assistants~\cite{zhang2015snap}. However, despite these technological strides, the benchmarks used to evaluate these capabilities remain surprisingly stagnant, often failing to reflect the complexity of real-world dining scenarios.

Existing food datasets, such as Food-101 \cite{bossard2014food} and UEC-Food \cite{matsuda2012recognition}, have largely driven progress in visual recognition. Yet, they suffer from four critical limitations when evaluated against the capabilities of modern VLMs. 
\textbf{First, tasks are overly simplistic.} Most benchmarks focus solely on coarse-grained classification, neglecting deeper reasoning capabilities such as nutritional quantification or culinary analysis. 
\textbf{Second, single-view limitation.} Traditional datasets typically treat food recognition as a single-image problem. In contrast, real-world user behavior involves capturing multiple angles to understand portion size and ingredients fully. 
\textbf{Third, lack of fine-grained discrimination.} Distractors in existing multiple-choice evaluations are often randomly sampled, allowing models to rely on superficial semantic priors rather than genuine visual understanding. \textbf{Fourth, inaccurate nutritional annotations.} Existing nutrition estimation datasets such as Recipe1M+~\cite{marin2019recipe1m+} suffer from low image quality
% and discrepancies between web-scraped images and actual nutritional metadata
, while Nutrition5K~\cite{thames2021nutrition5k} and FastFood~\cite{qi2025advancing} focus narrowly on standardized cafeteria or fast-food chain restaurant settings, limiting food diversity and real-world applicability.

To bridge this gap, we introduce \textbf{DiningBench}, a hierarchical, multi-view benchmark meticulously designed to evaluate VLMs across three levels of cognitive complexity: \textit{Fine-Grained Classification}, \textit{Nutrition Estimation}, and \textit{Visual Question Answering (VQA)}. 
Unlike previous efforts, DiningBench is constructed from rich, user-generated content sourced from distinct restaurants, ensuring a high degree of visual and semantic challenge.

DiningBench distinguishes itself through several key contributions:

\begin{itemize}[leftmargin=*, nosep]
\item \textbf{Hierarchical Task Design.} We propose a structured evaluation pipeline that moves from identification (Classification) to quantification (Nutrition Estimation) and finally to high-level reasoning (VQA). This tests not just what the model sees, but what it understands about volume, composition, and dietary implications.

\item \textbf{Multi-View Consistency.} DiningBench provides an average of 5.27 images per dish from different users and angles. This enables the study of multi-view information fusion.
% , simulating a realistic scenario where an agent must aggregate visual evidence to form a comprehensive conclusion.

\item \textbf{Fine-Grained Hard Discrimination.} By sourcing distractor options from the same merchant's menu within the same category, we construct a ``Hard'' classification setting. For instance, distinguishing a \textit{Smoked Salmon Salad} from a \textit{Fresh Salmon Avocado Salad} requires the model to identify subtle visual cues rather than relying on category-level differences.

\item \textbf{Comprehensive and High-Fidelity Nutrition Alignment.}  
% Addressing the critical gaps in both fidelity and comprehensiveness within existing datasets, we leveraged merchant-provided metadata alongside high-quality merchant and user-generated images, ensuring both comprehensiveness and accuracy.
We addressed existing dataset limitations by integrating merchant metadata with high-quality images, ensuring comprehensive and accurate coverage.

\item \textbf{Modern Construction Pipeline.} We introduce a rigorous AI-assisted data curation pipeline, leveraging state-of-the-art models (Qwen-2.5-VL, Gemini-3-Pro-Preview) for image quality assessment, reference matching, and nutrition inference, setting a new standard for efficient and high-quality dataset construction.
\end{itemize}

% Through extensive experiments, we demonstrate that while current SOTA MLLMs achieve impressive results on general benchmarks, they face significant challenges on DiningBench, particularly in fine-grained discrimination and precise nutritional regression. We hope DiningBench serves as a catalyst for developing next-generation food computing agents.

Our contributions are threefold:

\MakeUppercase{\romannumeral 1}. We construct DiningBench\footnote{\url{https://huggingface.co/datasets/meituan/DiningBench}.}, a comprehensive benchmark featuring hierarchical tasks and multi-view images, specifically designed to evaluate VLMs' capabilities in real-world food understanding scenarios.

\MakeUppercase{\romannumeral 2}. We conduct extensive evaluations on 29 state-of-the-art open-source and proprietary models, demonstrating significant performance gaps and revealing that current VLMs struggle with fine-grained visual discrimination and precise nutritional reasoning.
% , highlighting substantial room for improvement.

\MakeUppercase{\romannumeral 3}. We perform in-depth analysis, exploring the impact of multi-image input and Chain-of-Thought reasoning on model performance, and systematically analyzing the primary failure modes to provide insights for future research directions.

\begin{figure*}
 \centering
\includegraphics[width=\textwidth]{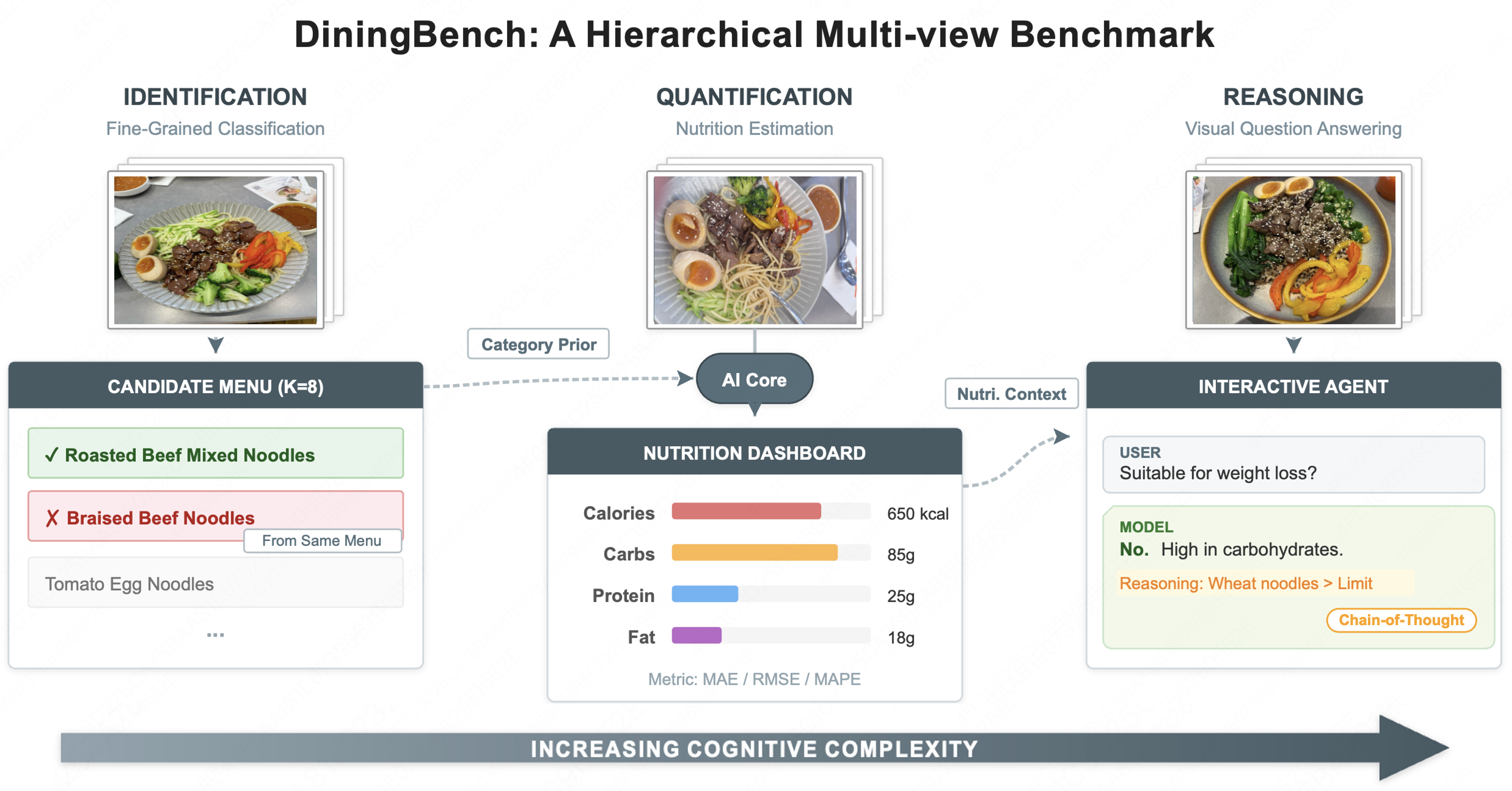}
\caption{Overview of the DiningBench Framework. The benchmark evaluates VLMs across a hierarchy of cognitive complexity: (1) Identification (Fine-Grained Classification with hard negatives), (2) Quantification (Nutrition Estimation), and (3) Reasoning (Visual Question Answering). The pipeline utilizes multi-view imagery to assess fine-grained visual understanding.}
\label{fig:main}
\vspace{-10pt}
\end{figure*}

\section{Related Work}

\subsection{General Multimodal Benchmarks}
Recent multimodal benchmarks have increasingly emphasized comprehensive evaluation across heterogeneous capabilities. VQA~\cite{antol2015vqa} and GQA~\cite{ainslie2023gqa} assess fine-grained perception 
and compositional reasoning
in static imagery, while broader benchmarks~\cite{liu2024mmbench,fu2025mme,  li2023seed, chen2024we}
% including MMBench~\cite{liu2024mmbench}, MME~\cite{fu2025mme}, SEED-Bench~\cite{li2023seed} and MMStar~\cite{chen2024we} 
evaluate cross-modal alignment, instruction following, and factual grounding across diverse domains.
Reasoning-oriented extensions introduce symbolic reasoning and quantitative problem solving under multimodal contexts~\cite{lu2023mathvista,masry2022chartqa,wang2024measuring}, and hallucination-specific evaluations investigate faithfulness and grounding reliability in open-ended generation~\cite{li2023evaluating, guan2024hallusionbench, li2025vidhalluc}. Meanwhile, emerging video benchmarks such as NExT-QA~\cite{xiao2021next} and MVBench~\cite{li2024mvbench} extend these objectives to dynamic scenes  and long-horizon multimodal understanding. However, these general-purpose benchmarks rarely capture the unique challenges of the food domain, where multimodal understanding involves fine-grained category discrimination, subtle visual attributes, strong domain knowledge dependencies, and higher risks of semantic and factual hallucination in health-related reasoning.
\subsection{Food Benchmarks}

% The landscape of food computing has been shaped by the evolution of benchmarks ranging from fundamental visual recognition to complex multimodal reasoning. 
Foundational datasets such as Food-101~\cite{bossard2014food}, UEC-Food~\cite{arslan2021fine}, VIREO Food-172~\cite{chen2016deep}, and ISIA Food-500~\cite{min2020isia} established robust baselines for food categorization, while Food2K~\cite{min2023large} further scaled visual feature learning through massive categorical coverage. To bridge the gap between visual appearance and procedural knowledge, cross-modal corpora like Recipe1M+~\cite{marin2019recipe1m+} and RecipeQA~\cite{yagcioglu2018recipeqa} were introduced to facilitate image-to-recipe retrieval and comprehension. In the domain of quantification, datasets such as Nutrition5k~\cite{thames2021nutrition5k} and FastFood~\cite{qi2025advancing} leverage VLMs to estimate caloric and nutrient content; however, these resources are often constrained by standardized cafeteria settings or noisy web-scraped metadata, limiting their generalization to diverse real-world dining scenarios. Furthermore, while recent benchmarks like FoodieQA~\cite{li2024foodieqa} and IndiFoodVQ~\cite{agarwal2024indifoodvqa} have pioneered culturally aware reasoning, they typically treat logic in culture. DiningBench distinguishes itself by unifying these dimensions into a hierarchical evaluation, progressing from identification via context-aware hard negatives to precise nutrition estimation and high-level reasoning, supported by multi-view consistency.

\section{DiningBench}

\subsection{Task Definitions}
To comprehensively evaluate VLMs in the food domain, we design a hierarchical benchmark comprising three distinct tasks: Fine-Grained Classification, Nutrition Estimation, and Visual Question Answering (VQA). All prompt formulation and dataset cases are provided in the Appendix~\ref{app:prompts} and Appendix~\ref{app:dataset_case}.

\textbf{Fine-Grained Classification.}
This task assesses the model's ability to distinguish between visually similar food categories. Formally, given an input image set $\mathcal{I}$ containing one or more images of a dish, the model is presented with a candidate set $\mathcal{C} = \{c_1, c_2, \dots, c_K\}$ consisting of $K=8$ options. 
The set $\mathcal{C}$ includes one ground-truth label $y$ and $K-1$ hard distractors (e.g., \textit{Smoked Salmon Salad} vs. \textit{Salmon Avocado Salad}). The goal is to choose the correct option.

\textbf{Nutrition Estimation.}
This task evaluates the model's capability to quantify nutritional content purely from visual cues. For a given input image set $\mathcal{I}$, the model must predict a nutrition vector $\mathbf{v} \in \mathbb{R}^4$, representing \textit{Calories}, \textit{Carbohydrates}, \textit{Protein}, and \textit{Fat}. Unlike classification, this is a regression problem where the model must estimate continuous values based on portion size and ingredients inferred from $\mathcal{I}$.

\begin{figure*}
 \centering
\includegraphics[width=\textwidth]{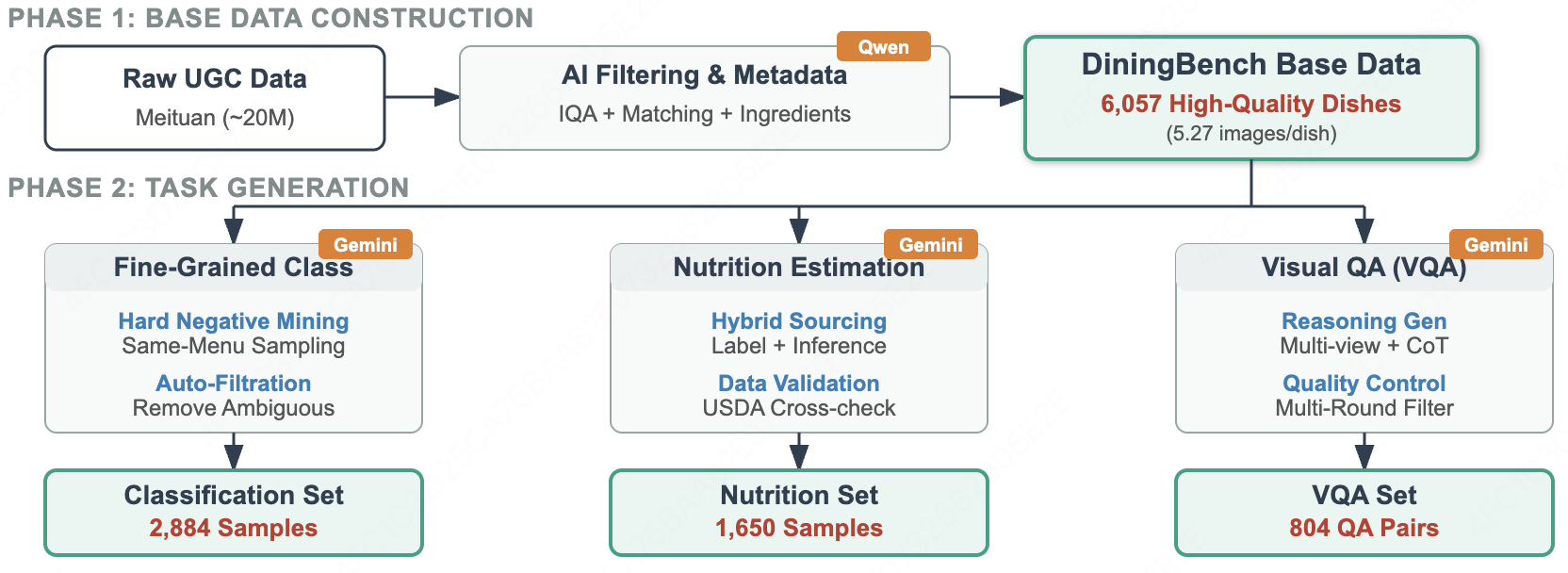}
\caption{DiningBench Data Construction Pipeline. The process is divided into two phases: (1) Base Data Construction, involving the filtration of raw user-generated content (UGC); and (2) Task Generation, utilizing AI-assisted pipelines (with Gemini-3-Pro-Preview) for hard negative mining, nutrition inference, and VQA generation, followed by rigorous human verification.}
\label{fig:pipeline}
\vspace{-10pt}
\end{figure*}

\textbf{Visual Question Answering (VQA).}
The VQA task probes higher-order reasoning capabilities. Given an image set $\mathcal{I}$ and a natural language question $q$, the model generates a textual response $a$. The questions cover complex dimensions including culinary techniques, dietary advice, multi-image comparative analysis, and counterfactual reasoning.

\subsection{Dataset Construction Pipeline}
\label{sec:dataset_construction}

This section details the acquisition of the Base Data and the rigorous pipeline employed to construct the task-specific datasets, as shown in Figure~\ref{fig:pipeline}.

\subsubsection{Base Data Acquisition}

The raw images and metadata were sourced from \textit{Meituan}\footnote{\url{https://www.meituan.com/}.}, China’s preeminent local life service platform, drawing upon its vast repository of authentic, multimodal dining content. This dataset encapsulates professional merchant-provided reference images alongside diverse, user-generated photos captured from varying perspectives, supplemented by textual attributes such as dish names, portion sizes, and descriptions.

% All raw images and food metadata were sourced from \textit{Dianping}, a user-generated content platform where users share photos of food items from various restaurants. The metadata encompasses merchant-provided reference images, user-generated images captured from diverse perspectives by different users, along with dish names, portion sizes, categories, and descriptions.

We implemented a rigorous multi-stage filtering pipeline to curate high-quality data. We began with approximately 20M user-generated images from the platform. For quality and consistency filtering, we employed knowledge distillation from GPT-4 to train two specialized discriminators based on Qwen-2.5-VL-7B: (1) an Image Quality Assessment model to evaluate visual quality, and (2) a Reference-Matching model to verify consistency between user photos and merchant reference images. Application of these models reduced the dataset to 685k images. Subsequently, images were grouped by dish, and dishes with fewer than three user photos were excluded through frequency thresholding, resulting in 90k distinct dishes. We then validated reference image quality, retaining 41k dishes with high-quality merchant reference images. For metadata enrichment, we selected dishes containing detailed ingredient lists in their descriptions, yielding 15k candidates. Finally, following category-based deduplication and balancing across cuisine origins, a manual quality check produced a Base Data comprising 6,057 high-quality, well-balanced dish entries, each accompanied by sufficient multi-view user photos.

\subsubsection{Fine-Grained Classification Dataset}
To evaluate fine-grained discriminative capabilities, we constructed a set of highly challenging negative samples. For each target dish, we employed Gemini-3-Pro-Preview to select seven visually or semantically similar items from the same category within the same merchant's menu. Sourcing candidates from the same menu category inherently ensures a high degree of similarity among items, thereby guaranteeing the distractiveness of the samples.
The candidate data then underwent a two-pass automated filtration process using Gemini-3-Pro-Preview and Gemini-2.5-Pro sequentially. We excluded overly ambiguous samples where image resolution or visual features were insufficient to uniquely identify the ground truth, and removed trivial samples where the distractors were too easily distinguishable from the target. The specific prompts used for item selection and filtration are detailed in the Appendix~\ref{app:prompts}. Following a final round of human verification for each filtered sample, the resulting dataset comprises 2,884 samples.

\begin{table*}[htbp]
  \centering

  \resizebox{0.75\linewidth}{!}{
    \begin{tabular}{lcccc}
    \toprule
    \textbf{Name} & \textbf{Dish Count} & \textbf{Image Count} & \textbf{Avg Images Per Dish} & \textbf{Dish Categories} \\
    \midrule
    \textbf{All} & 3021  & 15928 & 5.27  & 2060 \\
    \textbf{Fine-Grained Classification} & 2884  & 15330 & 5.32  & 1977 \\
    \textbf{Nutrition Estimation} & 1650  & 8856  & 5.37  & 1247 \\
    \textbf{Visual Question Answering} & 804   & 839   & 1.04  & 696 \\
    \bottomrule
    \end{tabular}%
    }
      \caption{DiningBench Dataset Statistics. A summary of the total number of dishes, images, and categories across the three subsets: Fine-Grained Classification, Nutrition Estimation, and Visual Question Answering (VQA).}
  \label{tab:dataset_statics}%
  \vspace{-10pt}
\end{table*}%

\begin{figure*}
 \centering
\includegraphics[width=\textwidth]{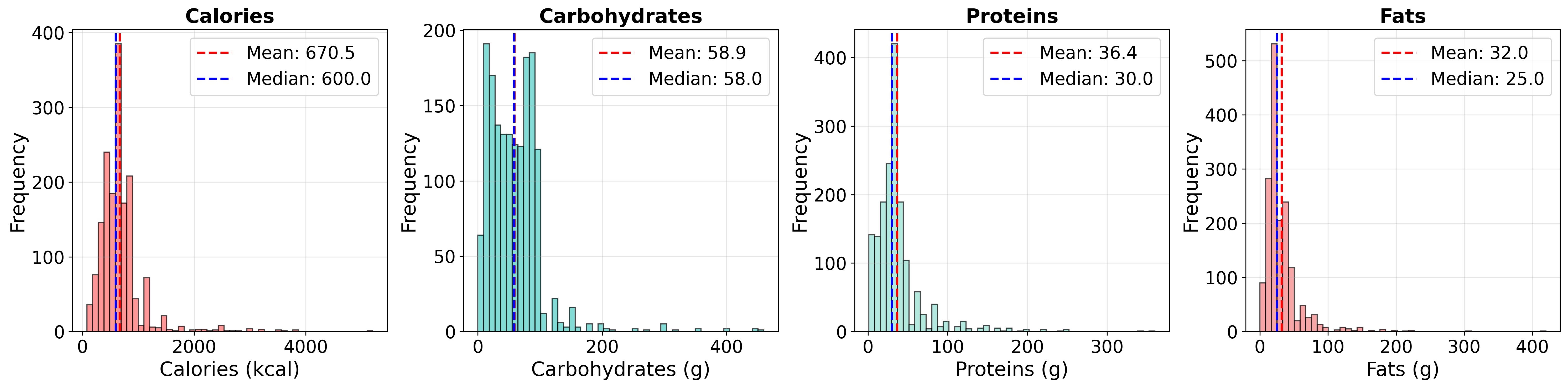}
\caption{
Distribution of Nutritional Values. 
Histograms illustrating the frequency distribution of Calories (kcal), Carbohydrates (g), Proteins (g), and Fats (g) across the DiningBench dataset, demonstrating a diverse range of nutritional profiles.}
\label{fig:nutrition_histograms}
\vspace{-10pt}
\end{figure*}

\subsubsection{Nutrition Estimation Dataset}
Ground truth nutrition data was obtained through two complementary methods. (1) Direct Extraction: For dishes where merchants explicitly provided nutritional information, we directly utilized the available data. (2) LLM-Assisted Estimation: For dishes lacking explicit nutritional labels but accompanied by detailed ingredient lists and corresponding portions, we employed Gemini-3-Pro-Preview. The model was prompted with the food image, ingredient composition, and portion sizes to generate nutritional estimates. The combination of a powerful language model and comprehensive source materials helped ensure reasonable data reliability. The specific prompts used are detailed in the Appendix~\ref{app:prompts}. To further validate the accuracy, we cross-referenced all generated estimates with the USDA FoodData Central\footnote{\url{https://fdc.nal.usda.gov/}.} database and conducted systematic manual verification. The final curated dataset comprises 1,650 samples.

\subsubsection{Visual Question Answering Dataset}
The VQA dataset was constructed in two batches to cover diverse reasoning types. Multi-Image Reasoning samples were created using dishes with at least two distinct images, requiring information synthesis across multiple views. Single-Image Reasoning samples focus on Cuisine Technique, Dietary Suggestion, and Counterfactual Reasoning based on metadata and visual content. We employed Gemini-3-Pro-Preview to generate these questions, with detailed prompts provided in the Table~\ref{tab:culinary_vqa_prompt} in Appendix~\ref{app:prompts}. CoT reasoning was enforced for all samples. Following question deduplication and two-round LLM filtering—which ensured answer uniqueness, clarity, reasoning correctness, appropriate difficulty levels, and relevance to food and images—along with manual verification, 804 high-quality samples were retained.

\subsection{Evaluation Metrics}

% We evaluate performance using standard Accuracy (Acc), defined as the ratio of correctly predicted options to the total number of samples.

% \textbf{Nutrition Estimation}
% We evaluate the regression performance for each of the four nutritional components: Calories, Carbohydrates, Protein, and Fat. For each component $k$, we calculate the Mean Absolute Error (MAE), Root Mean Square Error (RMSE), and Mean Absolute Percentage Error (MAPE). The final metric is the average across all four components. The MAPE for component $k$ is defined as:
% \begin{equation}
%     \text{MAPE}_k = \frac{1}{N} \sum_{i=1}^{N} \left| \frac{v_{i,k} - \hat{v}_{i,k}}{v_{i,k}} \right| 
% \end{equation}

% \textbf{Visual Question Answering}
% Given the open-ended nature of VQA, exact string matching is insufficient. We adopt an \textit{LLM-as-a-Judge} paradigm. An evaluator LLM compares the model's predicted answer $\hat{a}$ with the standard ground truth $a$ to determine semantic consistency and factual correctness. The performance is reported as Accuracy (Acc) based on the judge's binary verdict.

% To evaluate the model's performance comprehensively, we employ metrics tailored to the specific nature of each task. 
For Fine-Grained Classification, we utilize standard Accuracy (Acc), defined as the ratio of correctly predicted options. 
% In the task of
For Nutrition Estimation, we assess the regression performance for Calories, Carbohydrates, Protein, and Fat using Mean Absolute Error (MAE), Root Mean Square Error (RMSE), and Mean Absolute Percentage Error (MAPE~\cite{de2016mean}). Specifically, the MAPE for a component $k$ is formulated as follows:
\begin{equation}
    \text{MAPE}_k = \frac{1}{N} \sum_{i=1}^{N} \left| \frac{v_{i,k} - \hat{v}_{i,k}}{v_{i,k}} \right|,
\end{equation}
where $N$ is the total number of samples, $v_{i,k}$ denotes the ground truth value of the $k$-th nutritional component for the $i$-th sample, and $\hat{v}_{i,k}$ represents the corresponding predicted value.
The final metric is obtained by averaging these values across all four components. Lastly, for Visual Question Answering, we address the limitations of exact string matching by adopting an \textit{LLM-as-a-Judge} paradigm, where an evaluator LLM assesses the semantic consistency and factual correctness of the predicted answer $\hat{a}$ against the ground truth $a$, reporting the final performance as Accuracy based on the judge's binary verdict.

\begin{table*}[tbp!]
  \centering
  \scriptsize
  \setlength{\tabcolsep}{3.5pt} 
  \renewcommand{\arraystretch}{1.1} 
  \begin{adjustbox}{width=\textwidth,center}
    \begin{tabularx}{\textwidth}{
      l
      >{\raggedleft\arraybackslash}X  % Classification (Right aligned)
      *{7}{>{\raggedleft\arraybackslash}X}  % Nutrition (Right aligned)
      >{\raggedleft\arraybackslash}X  % VQA (Right aligned)
    }
    \toprule
    \multirow{2}{*}{\textbf{Models}} 
      & \multicolumn{1}{c}{\cellcolor{cyan!10}\textbf{Class.}} 
      & \multicolumn{7}{c}{\cellcolor{orange!10}\textbf{Nutrition Estimation}} 
      & \multicolumn{1}{c}{\cellcolor{purple!10}\textbf{VQA}} \\
    \cmidrule(lr){2-2} \cmidrule(lr){3-9} \cmidrule(lr){10-10}
      
      & \multicolumn{1}{c}{\tiny \textbf{ACC} $\uparrow$}
      & \multicolumn{1}{c}{\tiny Cal MAPE $\downarrow$} 
      & \multicolumn{1}{c}{\tiny Prot MAPE $\downarrow$} 
      & \multicolumn{1}{c}{\tiny Carbs MAPE $\downarrow$} 
      & \multicolumn{1}{c}{\tiny Fat MAPE $\downarrow$} 
      & \multicolumn{1}{c}{\tiny Avg MAE $\downarrow$} 
      & \multicolumn{1}{c}{\tiny Avg RMSE $\downarrow$} 
      & \multicolumn{1}{c}{\tiny Avg MAPE $\downarrow$} 
      & \multicolumn{1}{c}{\tiny \textbf{ACC} $\uparrow$} \\
    \midrule

    % ================= Proprietary Models =================
    \rowcolor{gray!10}\multicolumn{10}{l}{\emph{\textbf{Proprietary Models}}} \\
    Claude-Sonnet-4.5       & 0.5440 & 35.01 & 33.15 & 51.52 & 50.82 & 91.14 & 140.82 & 42.62 & 0.8358 \\
    Gemini-2.5-Flash        & 0.7101 & 30.21 & 49.02 & 42.62 & 44.56 & 74.33 & 122.30 & 41.60 & 0.7836 \\
    Gemini-2.5-Pro          & 0.7351 & 28.61 & 39.02 & 43.01 & 42.19 & 68.90 & 124.19 & 38.21 & \underline{0.8993} \\
    Gemini-3-Flash-Preview  & \textbf{0.8183} & \underline{20.57} & \underline{25.23} & \underline{27.94} & \underline{27.09} & \underline{55.54} & \textbf{103.52} & \underline{25.21} & 0.8856 \\
    Gemini-3-Pro-Preview    & \underline{0.8155} & \textbf{19.99} & \textbf{23.88} & \textbf{27.40} & \textbf{26.53} & \textbf{55.11} & \underline{105.06} & \textbf{24.45} & \textbf{0.9042} \\
    GPT-4.1                 & 0.6859 & 36.88 & 32.42 & 35.81 & 49.29 & 94.27 & 143.31 & 38.60 & 0.8433 \\
    GPT-4o                  & 0.6526 & 41.67 & 36.03 & 40.19 & 51.82 & 105.65 & 154.32 & 42.43 & 0.8060 \\
    GPT-4o-mini             & 0.5274 & 41.17 & 35.08 & 45.03 & 55.05 & 106.19 & 156.54 & 44.08 & 0.7139 \\
    GPT-5                   & 0.7018 & 25.14 & 33.40 & 36.63 & 33.49 & 68.38 & 120.34 & 32.17 & 0.8694 \\
    O4-mini                 & 0.6481 & 27.68 & 32.96 & 37.38 & 37.80 & 74.70 & 125.28 & 33.95 & 0.8035 \\

    \midrule
    % ================= Open-source Models =================
    \rowcolor{gray!10}\multicolumn{10}{l}{\emph{\textbf{Open-Source Models}}} \\
    Gemma-3-12B-it          & 0.4861 & 27.95 & 40.28 & 63.24 & 41.12 & 71.68 & 114.91 & 43.15 & 0.6182 \\
    InternVL-3.5-14B        & 0.4955 & 45.18 & 39.16 & 60.30 & 57.86 & 114.96 & 160.90 & 50.62 & 0.6779 \\
    InternVL-3.5-30B-A3B    & 0.4927 & 38.81 & 38.03 & 48.65 & 53.56 & 99.32 & 146.74 & 44.76 & 0.7040 \\
    InternVL-3.5-38B        & 0.5420 & 41.53 & 35.59 & 49.79 & 57.60 & 107.01 & 153.77 & 46.13 & 0.7251 \\
    InternVL-3.5-4B         & 0.4376 & 45.13 & 38.85 & 60.54 & 60.11 & 114.23 & 159.21 & 51.16 & 0.6455 \\
    InternVL-3.5-8B         & 0.4532 & 38.16 & 38.86 & 72.59 & 51.23 & 93.50 & 133.48 & 50.21 & 0.6480 \\
    Keye-VL-1.5-8B          & 0.5555 & 38.21 & 40.46 & 56.41 & 52.09 & 95.04 & 140.43 & 46.79 & 0.6580 \\
    Mimo-VL-7B-RL           & 0.5638 & 42.90 & 39.89 & 48.24 & 58.82 & 109.31 & 156.91 & 47.46 & 0.7500 \\
    MiniCPM-V-4.5           & 0.5558 & 34.79 & 40.47 & 53.73 & 49.69 & 86.63 & 127.14 & 44.67 & 0.5821 \\
    Qwen-2.5-VL-32B-Instruct   & 0.6117 & 32.66 & 32.73 & 43.92 & 50.05 & 86.27 & 132.52 & 39.84 & 0.7114 \\
    Qwen-2.5-VL-3B-Instruct    & 0.5149 & 37.67 & 37.46 & 64.53 & 57.57 & 100.20 & 148.24 & 49.31 & 0.4764 \\
    Qwen-2.5-VL-72B-Instruct   & 0.6529 & 35.88 & 33.38 & 41.70 & 51.27 & 94.22 & 143.23 & 40.56 & 0.7662 \\
    Qwen-2.5-VL-7B-Instruct    & 0.6085 & 40.67 & 34.67 & 51.03 & 56.55 & 106.09 & 151.55 & 45.73 & 0.6169 \\
    Qwen-3-VL-30B-A3B-Instruct  & 0.6543 & 32.49 & 32.01 & 41.53 & 43.37 & 85.69 & 137.63 & 37.35 & 0.8060 \\
    Qwen-3-VL-30B-A3B-Think & 0.6134 & 32.33 & 36.77 & 48.12 & 43.66 & 82.80 & 128.74 & 40.22 & 0.7413 \\
    Qwen-3-VL-4B-Instruct   & 0.6006 & 40.37 & 36.00 & 49.28 & 44.55 & 101.88 & 146.82 & 42.55 & 0.7077 \\
    Qwen-3-VL-4B-Thinking   & 0.5742 & 52.91 & 49.96 & 62.40 & 62.26 & 125.81 & 171.54 & 56.88 & 0.6704 \\
    Qwen-3-VL-8B-Instruct   & 0.6415 & 31.06 & 33.87 & 43.99 & 48.02 & 78.90 & 123.41 & 39.24 & 0.7276 \\
    Qwen-3-VL-8B-Thinking   & 0.5898 & 50.99 & 40.04 & 52.76 & 63.88 & 123.39 & 167.89 & 51.92 & 0.6853 \\
    \bottomrule
    \end{tabularx}
  \end{adjustbox}
  \vspace{-0.5em}
  \caption{A comprehensive comparison of 29 proprietary and open-source VLMs across the three DiningBench tasks. The \textbf{best} and \underline{second-best} results are highlighted in bold and underlined, respectively.}
  \label{tab:main_results}
\vspace{-10pt}
\end{table*}

\subsection{Dataset Statistics}

We present a comprehensive statistical analysis of DiningBench, covering the overall dataset composition, nutritional value distributions, VQA task categories, and geographic diversity.

\textbf{Overall Composition.}
As summarized in Table~\ref{tab:dataset_statics}, DiningBench comprises 3,021 unique dishes spanning 2,060 distinct categories, supported by 15,928 high-quality images. The dataset features rich multi-view visual information, averaging 5.27 images per dish. The Fine-Grained Classification subset contains 2,884 dishes with 15,330 images (5.32 images/dish). The Nutrition Estimation subset includes 1,650 dishes with 8,856 images.
% (5.37 images/dish). 
% The Visual Question Answering subset consists of 804 QA pairs.

\textbf{Nutritional Value Distribution.}
To ensure robustness for the regression task, we analyze the distribution of four key nutritional components. As illustrated in Figure~\ref{fig:nutrition_histograms}, the dataset covers a wide range of nutritional profiles. Calories exhibit a mean of 670.5 kcal, encompassing both light meals and calorie-dense dishes. Carbohydrates, Proteins and Fats also show diverse distributions.

\textbf{VQA Task Diversity.}    
The VQA dataset encompasses diverse food-related tasks. As shown in the Table~\ref{tab:dataset_stats2}, Cuisine Technique (532 samples) and Dietary Suggestion (219 samples) constitute the primary focus, requiring models to identify cooking methods and provide practical health recommendations. Additionally, the dataset incorporates challenging reasoning tasks: Multi-Image Analysis (35 samples) evaluates cross-view information synthesis and Counterfactual Reasoning (18 samples) assesses the model's capacity to reason about hypothetical scenarios.

\textbf{Geographic and Cultural Diversity.}
DiningBench features broad international coverage, ensuring applicability across global food cultures. While Chinese cuisine forms a substantial foundation (2,086 dishes) due to data sourcing, the dataset includes significant international representation. As detailed in the Table~\ref{tab:dataset_stats2}, it contains Western (286), Worldwide (252), Asian excluding Chinese/Japanese (187), Japanese (118), Latin-American, and Indian cuisines. This distribution ensures  evaluation across both regional specialties and diverse global culinary traditions.

% Table~\ref{tab:dataset_stats2} provides a detailed breakdown of the dataset's geographic distribution and VQA task categories, highlighting the diversity of cuisines and reasoning types included in DiningBench.

\begin{table}[h]
\centering

\resizebox{\linewidth}{!}{
    \begin{tabular}{lrlr}
    \toprule
    \multicolumn{2}{c}{\textbf{Country Distribution}} & \multicolumn{2}{c}{\textbf{VQA Category Distribution}} \\
    \cmidrule(lr){1-2} \cmidrule(lr){3-4}
    \textbf{Region} & \textbf{Count} & \textbf{Category} & \textbf{Count} \\
    \midrule
    Chinese        & 2086 & Cuisine Technique         & 532 \\
    Western        & 286  & Dietary Suggestion        & 219 \\
    Worldwide      & 252  & Multi Image Analysis      & 35  \\
    Asian          & 187  & Counterfactual Reasoning  & 18  \\
    Japanese       & 118  &                           &     \\
    Latin-American & 48   &                           &     \\
    Indian         & 44   &                           &     \\
    \bottomrule
    \end{tabular}
}
\caption{A breakdown of the geographic distribution of cuisines and the categorization of VQA tasks.}
\label{tab:dataset_stats2} 
\end{table}

\section{Experiments}

We conduct comprehensive experiments and in-depth analyses to evaluate the value and utility of DiningBench, with supplementary experiments provided in Appendix~\ref{sec:supplementary_experiments}.

\subsection{Experimental Setup}
\noindent \textbf{Evaluated Models.} We evaluate a comprehensive set of 29 models, comprising 10 proprietary and 19 open-source models. The proprietary set includes the Claude, Gemini, and GPT series. The open-source lineup features the InternVL and Qwen-VL series, alongside other competitive models such as Gemma-3-12B-it, Keye-VL-1.5-8B, Mimo-VL-7B-RL, and MiniCPM-V-4.5. Detailed model specifications are provided in the Appendix~\ref{app:prompts}.

\noindent \textbf{Implementation Details.} For proprietary models, we utilize official APIs with the temperature set to 0 and a maximum context length of 16,384 tokens to ensure reproducibility. Open-source models are deployed using vLLM. 
% To accommodate varying parameter scales, 
Models under 8B are deployed on a single NVIDIA A100 GPU; models between 30B and 38B utilize two A100 GPUs; and the 72B model requires four A100 GPUs. Consistent inference parameters (temperature=0, max tokens=16,384) are maintained.
% across all deployments.

\subsection{Performance of VLMs on DiningBench}
Table~\ref{tab:main_results} summarizes the performance of 29 VLMs across the three tasks. The results demonstrate that DiningBench presents a rigorous challenge, identifying significant gaps in the fine-grained visual understanding capabilities of even the most advanced models.

\textbf{Fine-Grained Classification exposes perceptual limitations.} While Gemini-3-Flash-Preview achieves a leading accuracy of 81.83\%, other top-tier models exhibit notable difficulties with hard distractors. Specifically, GPT-4o and GPT-5 attain accuracies of only 65.26\% and 70.18\%, respectively. This performance disparity underscores the challenge of distinguishing visually similar dishes and validates the efficacy of our adversarial data construction pipeline. The inability of powerful models to consistently discriminate between subtle visual features highlights that fine-grained recognition remains a significant bottleneck.

\textbf{Nutrition Estimation remains an open challenge.} Quantifying nutritional content from visual cues proves to be the most demanding task. Even the state-of-the-art Gemini-3-Pro-Preview yields an Average MAPE of 24.45\%, reflecting a non-negligible margin of error. The challenge is more pronounced for GPT-4o, which suffers from a high error rate of 42.43\%. These findings suggest that current VLMs lack the precise volumetric reasoning and ingredient analysis capabilities necessary for accurate regression, marking this as a critical avenue for future research.

\textbf{Complex food-related reasoning requires better visual grounding.} Although models generally perform better on VQA, the task is far from saturated. Competitive models such as GPT-4o achieve 80.60\% accuracy, leaving room for improvement in handling complex food queries. 
% Notably, the performance of Claude-Sonnet-4.5 (83.58\% on VQA vs. 54.40\% on Classification) reveals a critical disconnect: while high-level semantic reasoning is robust, it is constrained by the precision of the underlying visual perception.

 No single model comprehensively solves the benchmark. The widespread struggle with nutrition estimation and the inconsistent performance in fine-grained classification demonstrate that food-domain multimodal understanding is far from solved. DiningBench thus serves as a valuable testbed for advancing visually precise and domain-aware VLMs.

\subsection{Impact of Multi-View Imagery}
\label{sec:multi_view}
\begin{figure*}
 \centering
\includegraphics[width=\textwidth]{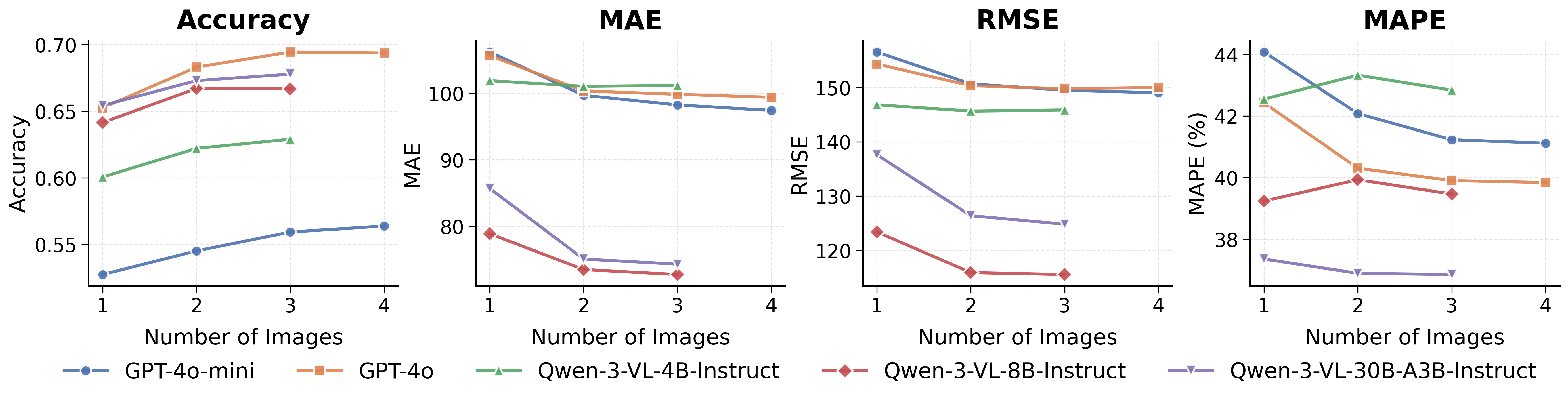}
\caption{Impact of Multi-View Inputs. Performance trends for Classification (Accuracy) and Nutrition Estimation (MAE, RMSE, MAPE) as the number of input images increases from 1 to 4. The results highlight the performance gains from view synthesis and the divergence between large-scale and smaller-scale models.}
\label{fig:multi-view}
\vspace{-10pt}
\end{figure*}

To investigate the efficacy of multi-view visual information, we evaluated representative models (GPT-4o and Qwen-3-VL series) on Fine-Grained Classification and Nutrition Estimation while varying the input image count from 1 to 4. Results are visualized in Figure~\ref{fig:multi-view}.

\textbf{Gains from Multi-View Synthesis.} Increasing the number of visual views generally enhances performance. Metrics such as Accuracy, MAE, and RMSE typically improve as the image count rises. The most significant performance leap occurs during the transition from a single view to two views ($1 \to 2$), indicating a substantial "capability jump" where complementary angles help resolve occlusion and ambiguity. However, marginal gains diminish with subsequent additions ($N > 2$), suggesting a saturation point in information utilization for current architectures.

\textbf{Performance Divergence Across Model Scales.} We observe a correlation between model scale and the ability to leverage multi-view data. Large-scale models, such as GPT-4o and Qwen-3-VL-30B-A3B, demonstrate consistent improvements. Conversely, smaller models exhibit instability, with MAPE fluctuating or degrading as more images are added. This phenomenon implies that for models with limited capacity, excessive visual tokens may act as noise or cause information overload rather than providing helpful context. This highlights effective multi-view fusion as an unresolved challenge, particularly for efficient, smaller-scale models.

\subsection{Effectiveness of CoT}

\begin{figure}
\centering
\includegraphics[width=\linewidth]{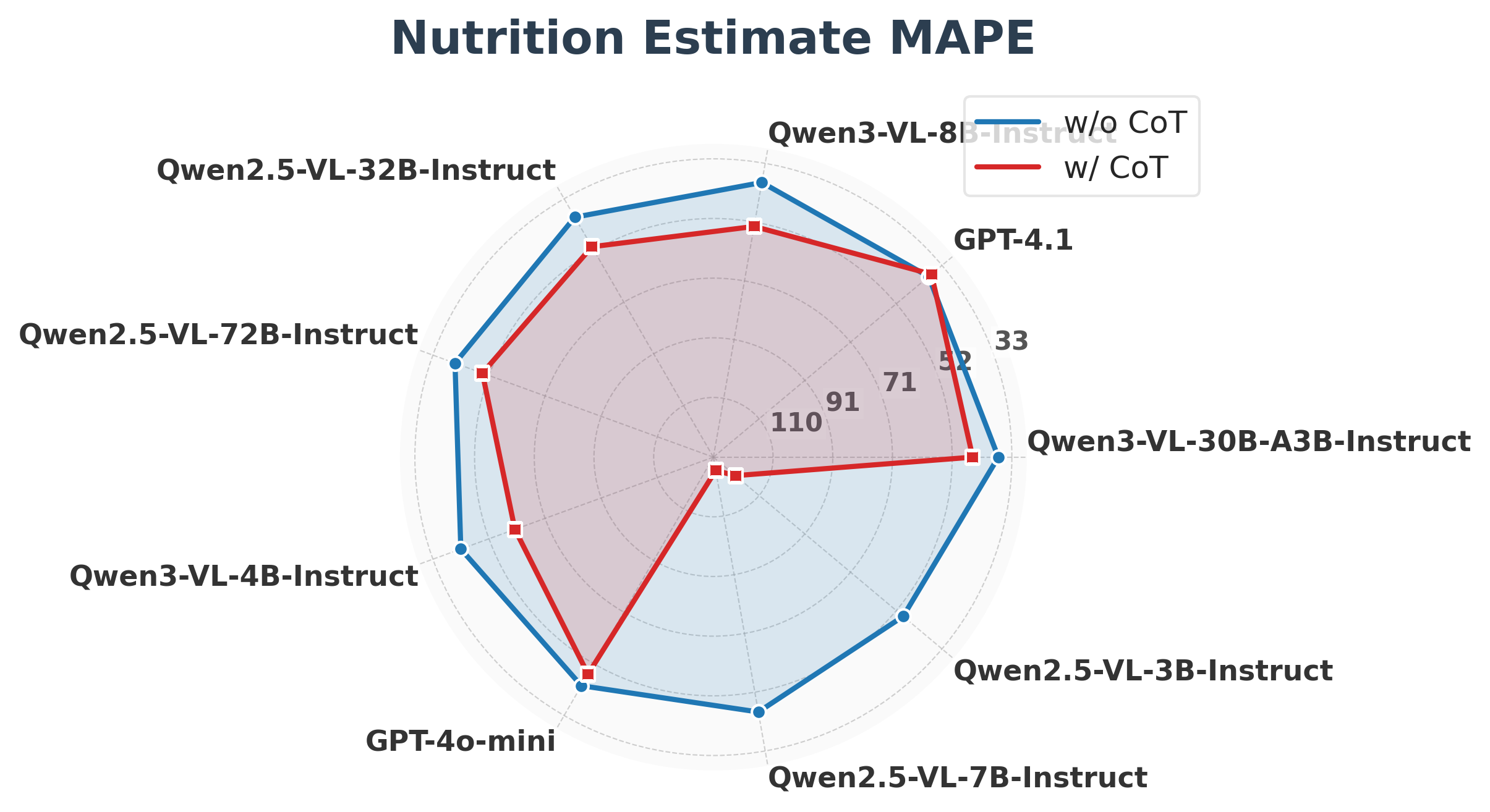}
\caption{Impact of CoT on Nutrition Estimation. A radar chart comparing the Mean Absolute Percentage Error (MAPE) of various models with and without CoT prompting. Higher values indicate higher error rates, showing that CoT often degrades performance for regression tasks in smaller models.}
\label{fig:nutrition_estimage_mape}
\vspace{-14pt}
\end{figure}

We conducted comprehensive experiments applying Chain-of-Thought (CoT) prompting across all tasks (templates provided in the Appendix~\ref{app:prompts}). Comparative visualizations (Figure~\ref{fig:nutrition_estimage_mape}, \ref{fig:classification_acc},  \ref{fig:vqa_accuraccy}) reveal that CoT is not universally beneficial for VLMs.

\textbf{CoT hinders direct visual perception tasks.} For tasks demanding precise visual discrimination and regression, specifically Nutrition Estimation, CoT often proved detrimental. As shown in Figure~\ref{fig:classification_acc}, most models experienced a decline in Classification Accuracy with CoT enabled. This negative impact is particularly severe in Nutrition Estimation (Figure~\ref{fig:nutrition_estimage_mape}), where smaller open-source models suffered a "performance collapse." Instead of refining predictions, the generated reasoning steps appear to introduce noise, drastically increasing MAPE. This suggests that explicit verbalization may decouple the final prediction from direct visual evidence, leading to hallucination or the over-rationalization of incorrect features.

\textbf{Mixed results on VQA.} In VQA, the efficacy of CoT is inconsistent. While select proprietary models achieved gains, others, including GPT-4o-mini and smaller Qwen variants, suffered performance degradation.

CoT is not a ``silver bullet'' for the DiningBench. Its effectiveness is heavily constrained by the model's fundamental visual grounding capabilities. When initial perception is flawed, CoT tends to amplify errors through a chain of incorrect reasoning rather than correcting them.

\subsection{Factors Contributing to Suboptimal Performance}
A qualitative error analysis reveals five primary factors hindering performance on DiningBench:

\textbf{Limited Fine-Grained Discriminability.} The most significant bottleneck is the lack of discriminative granularity. Our dataset's ``hard distractors'' (sharing ingredients/colors with the target) reveal that current VLMs often function as ``bag-of-features'' detectors. They identify dominant components but fail to perceive subtle distinctions in cutting styles or textures. A common error involves confusing \textit{Tomato Beef Pot} with \textit{Spicy Beef Pot} due to similar color tones, indicating a reliance on high-level semantics over low-level details.

\textbf{Parametric Knowledge Bias and Hallucination.} Models frequently rely on parametric knowledge priors rather than visual grounding. When encountering ambiguity or long-tail regional specialties, models often default to statistically probable dish names rather than the specific variant present in the image. For instance, models often misclassify \textit{Scallion Oil Chicken} as the more generic \textit{Roasted Chicken}, effectively ignoring contradictory visual evidence in favor of familiar text priors.

\textbf{Deficiencies in Spatial and Volumetric Reasoning.} High error rates in Nutrition Estimation stem from an inability to perform 2D-to-3D inference. Accurately predicting macronutrients requires estimating mass and volume relative to containers. Current models struggle with depth cues and scale, often treating appetizers and main courses as nutritionally equivalent if they share visual textures, indicating a lack of physical world understanding.

\textbf{Ineffective Multi-View Aggregation.} As noted in Section~\ref{sec:multi_view}, performance plateaus or declines when $N \ge 3$. Models struggle to synthesize complementary information or filter redundant features. Consequently, increased visual context often acts as noise, confusing the prediction rather than clarifying it.

\textbf{Inference Instability in Reasoning Models.} While ``Thinking'' models show promise, smaller-scale reasoning models exhibit instability, occasionally falling into ``infinite thinking loops''. We observed that visual uncertainty can trigger repetitive generation cycles where the model fails to converge on a conclusion, leading to valid but non-terminating reasoning steps.

\section{Additional Perspectives of DiningBench}
Beyond the evaluations presented in this work, DiningBench holds significant potential for broader research. Its high-quality, multi-view structure (averaging 5.27 images per dish) serves as a unique resource for 3D Reconstruction and Novel View Synthesis, introducing real-world challenges, such as complex occlusions and variable lighting, which are often absent in synthetic datasets. Furthermore, the alignment of professional reference images with user-generated content enables advanced research in Cross-Domain Retrieval and Conditional Image Generation.

\section{Ethical Considerations}
We strictly adhere to copyright, intellectual property, and privacy regulations. We have officially obtained explicit permission and copyright authorization from the data provider, Meituan, to utilize and distribute the base images and metadata for non-commercial research purposes. Consequently, DiningBench is legally compliant and is released under the CC BY-NC-ND 4.0 license. Furthermore, our dataset strictly excludes any Personally Identifiable Information (PII) and sensitive content. Prior to manual verification, all data underwent rigorous automated filtering to ensure the images exclusively depict safe, food-related content without any privacy-compromising elements.

\section{Conclusion}
We present DiningBench, a comprehensive benchmark designed to evaluate VLMs in the food domain through hierarchical tasks: Fine-Grained Classification, Nutrition Estimation, and Visual Question Answering. Leveraging high-quality and multi-view images, DiningBench exposes critical limitations in current VLMs, including insufficient fine-grained visual discrimination, deficient nutritional  reasoning, and ineffective multi-view fusion. Our extensive evaluation of 29 state-of-the-art models reveals substantial performance gaps, with even the strongest models struggling on nutrition quantification and subtle visual distinction. We hope DiningBench serves as a catalyst for advancing visually-grounded, domain-aware VLMs, ultimately contributing to more reliable AI systems for real-world food understanding and promoting healthier lifestyles and improved dietary outcomes worldwide.

\section{Limitations \& Potential Risks}

Despite the rigorous construction of DiningBench, several limitations remain. First, the dataset exhibits a cultural skew towards Chinese cuisine due to the sourcing platform, potentially affecting generalization across underrepresented global culinary traditions despite our efforts to include international dishes. Second, the reliance on LLM-assisted generation for distinct parts of the nutritional ground truth and distractor selection, although systematically verified by humans, may inevitably inherit latent biases or subtle inaccuracies. Regarding potential risks, the deployment of VLMs evaluated on this benchmark for real-world dietary guidance necessitates extreme caution. Errors in fine-grained classification or nutrition estimation could lead to incorrect health monitoring or allergen oversight, underscoring the critical need for human-in-the-loop oversight in practical health-related applications.

\section*{Acknowledgments}
This work is supported by Meituan. This work is also supported by the Public Computing Cloud, Renmin University of China and by fund for building worldclass universities (disciplines) of Renmin University of China.

% \nocite{Ando2005,andrew2007scalable,rasooli-tetrault-2015}

\bibliography{custom}

\clearpage

\appendix

\section{Dataset Case}
\label{app:dataset_case}

\begin{figure}
\centering
\includegraphics[width=\linewidth]{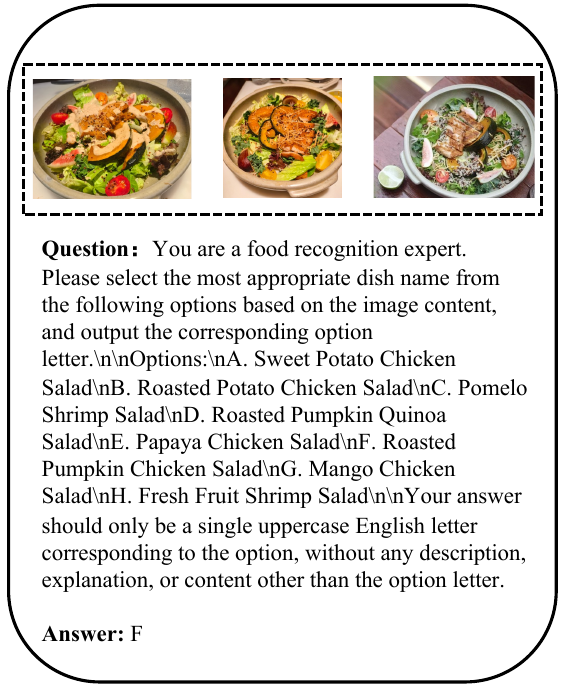}
\caption{Sample Case: Fine-Grained Classification. An example of the classification task where the model must identify the correct dish ("Roasted Pumpkin Chicken Salad") from a list of visually and semantically similar distractors sourced from the same menu.}
\label{fig:datataset1_case}
\end{figure}

\begin{figure}
\centering
\includegraphics[width=\linewidth]{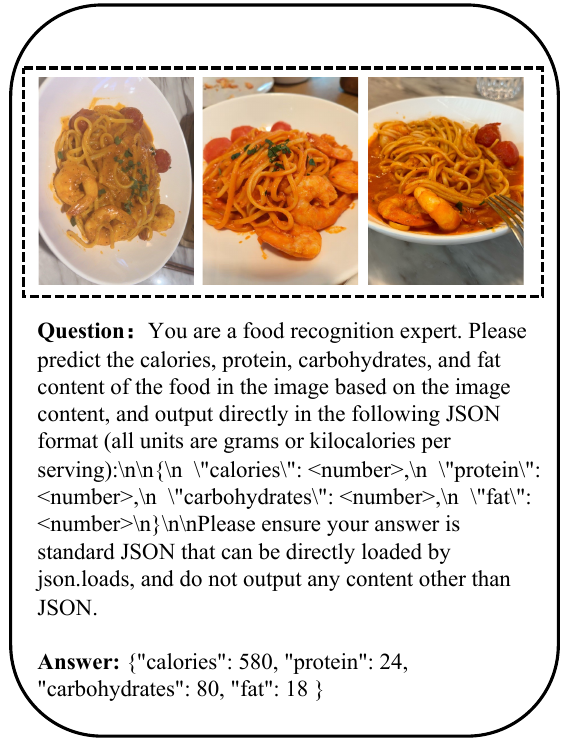}
\caption{Sample Case: Nutrition Estimation. An example showing the prompt and expected JSON output for quantifying Calories, Protein, Carbohydrates, and Fat based on visual inputs.}
\label{fig:datataset2_case}
\end{figure}

\begin{figure}
\centering
\includegraphics[width=\linewidth]{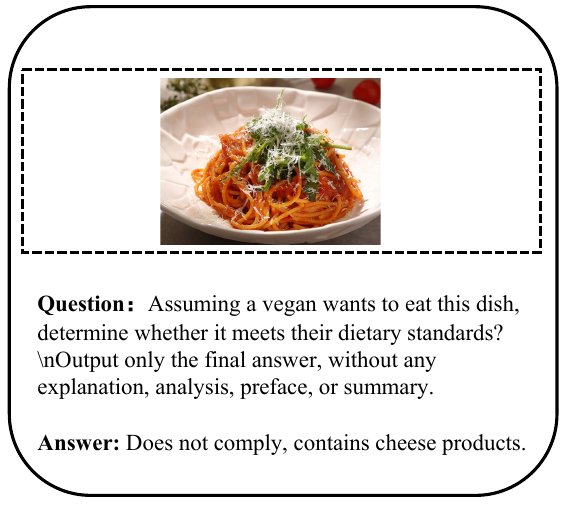}
\caption{Sample Case: Visual Question Answering. An example of a reasoning task where the model must determine if a dish meets specific dietary standards based on fine-grained visual cues like cheese crumbs.}
\label{fig:datataset3_case}
\end{figure}

We present specific examples for each task to illustrate the benchmark's difficulty and format. Figure~\ref{fig:datataset1_case} demonstrates a Fine-Grained Classification case with hard distractors. Figure~\ref{fig:datataset2_case} shows the input and output format for Nutrition Estimation. Figure~\ref{fig:datataset3_case} illustrates a Visual Question Answering (VQA) scenario requiring dietary reasoning.

\section{Supplementary Experiments}
\label{sec:supplementary_experiments}

\subsection{Baseline Details}
We conducted a comprehensive evaluation of 29 state-of-the-art VLMs, comprising 10 proprietary models and 19 open-source models. The baselines are categorized by model series below:

\paragraph{Gemini Series} We evaluate both the 2.5 series and the latest 3.0 preview series: Gemini-3-Flash-Preview, Gemini-3-Pro-Preview, Gemini-2.5-Flash, and Gemini-2.5-Pro~\cite{comanici2025gemini}.

\paragraph{GPT Series} This category includes current flagship models and advanced iterations: GPT-4o, GPT-4o-mini, GPT-4.1, GPT-5, and o4-mini~\cite{achiam2023gpt}.

\paragraph{Qwen-VL Series} We test a wide range of parameter scales from the Qwen family, including the latest Qwen-3 generation featuring ``Thinking'': Qwen-3-VL-4B (Instruct/Thinking), Qwen-3-VL-8B (Instruct/Thinking), Qwen-3-VL-30B-A3B (Instruct/Thinking), Qwen-2.5-VL-3B-Instruction, Qwen-2.5-VL-7B-Instruction, Qwen-2.5-VL-32B-Instruction, Qwen-2.5-VL-72B-Instruction~\cite{bai2025qwen2}.

\paragraph{InternVL Series} We evaluate the InternVL-3.5 series across various parameter: InternVL-3.5-4B, InternVL-3.5-8B, InternVL-3.5-14B, InternVL-3.5-30B-A3B, and InternVL-3.5-38B~\cite{wang2025internvl3}.

\paragraph{Other Models} This category includes other competitive proprietary and open-source models: Gemma-3-12B-it, Keye-VL-1.5-8B, Mimo-VL-7B-RL, MiniCPM-V-4.5~\cite{team2025gemma, yang2025kwai, xiaomi2025mimo, yu2025minicpm}.

\subsection{Supplementary for Effectiveness of Chain-of-Thought
(CoT) Prompting}

\begin{figure}
\centering
\includegraphics[width=\linewidth]{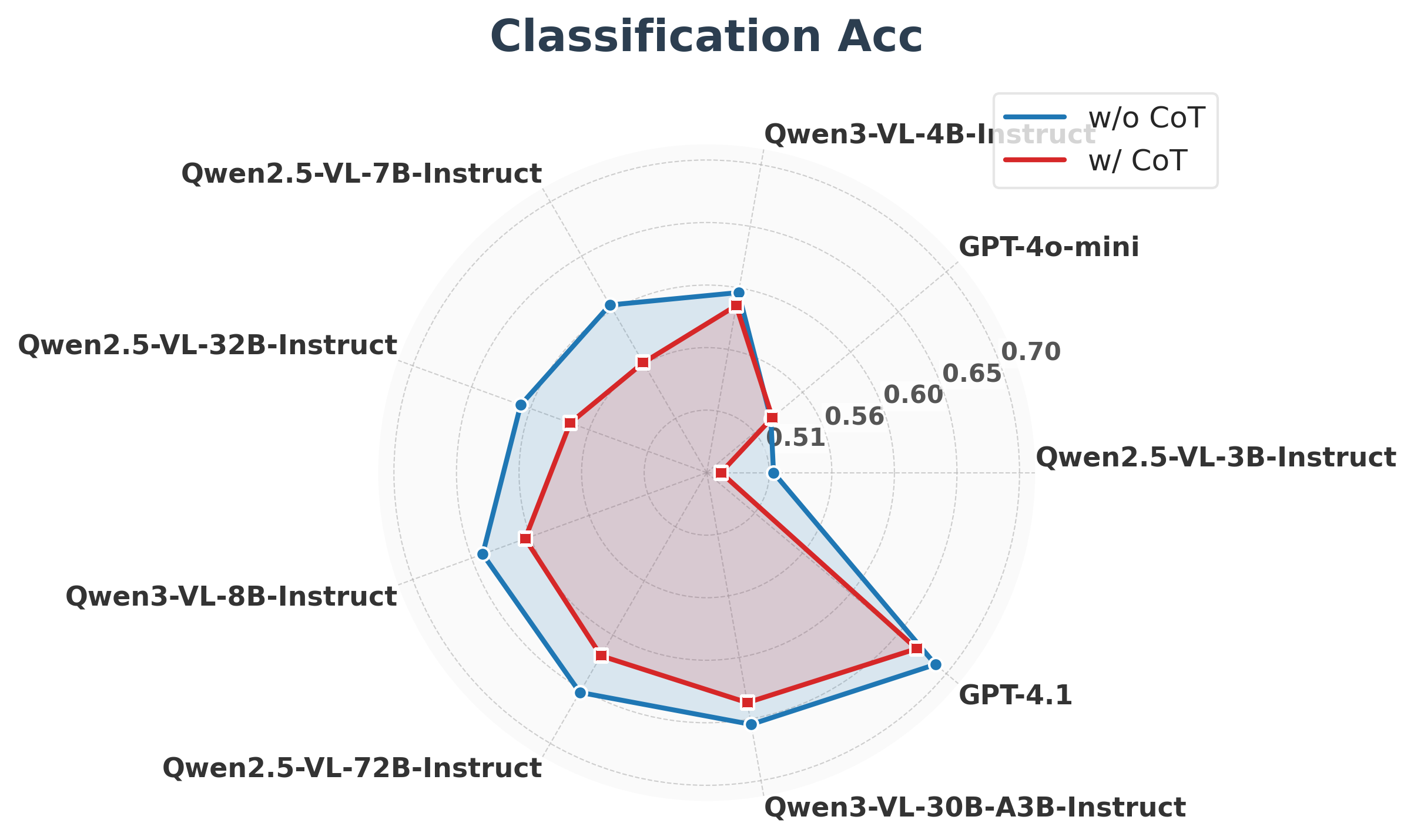}
\caption{Impact of Chain-of-Thought on Classification. A radar chart comparing the Classification Accuracy of various models with and without Chain-of-Thought (CoT) prompting, revealing that explicit reasoning steps can hinder direct visual discrimination in some models.}
\label{fig:classification_acc}
\end{figure}

\begin{figure}
\centering
\includegraphics[width=\linewidth]{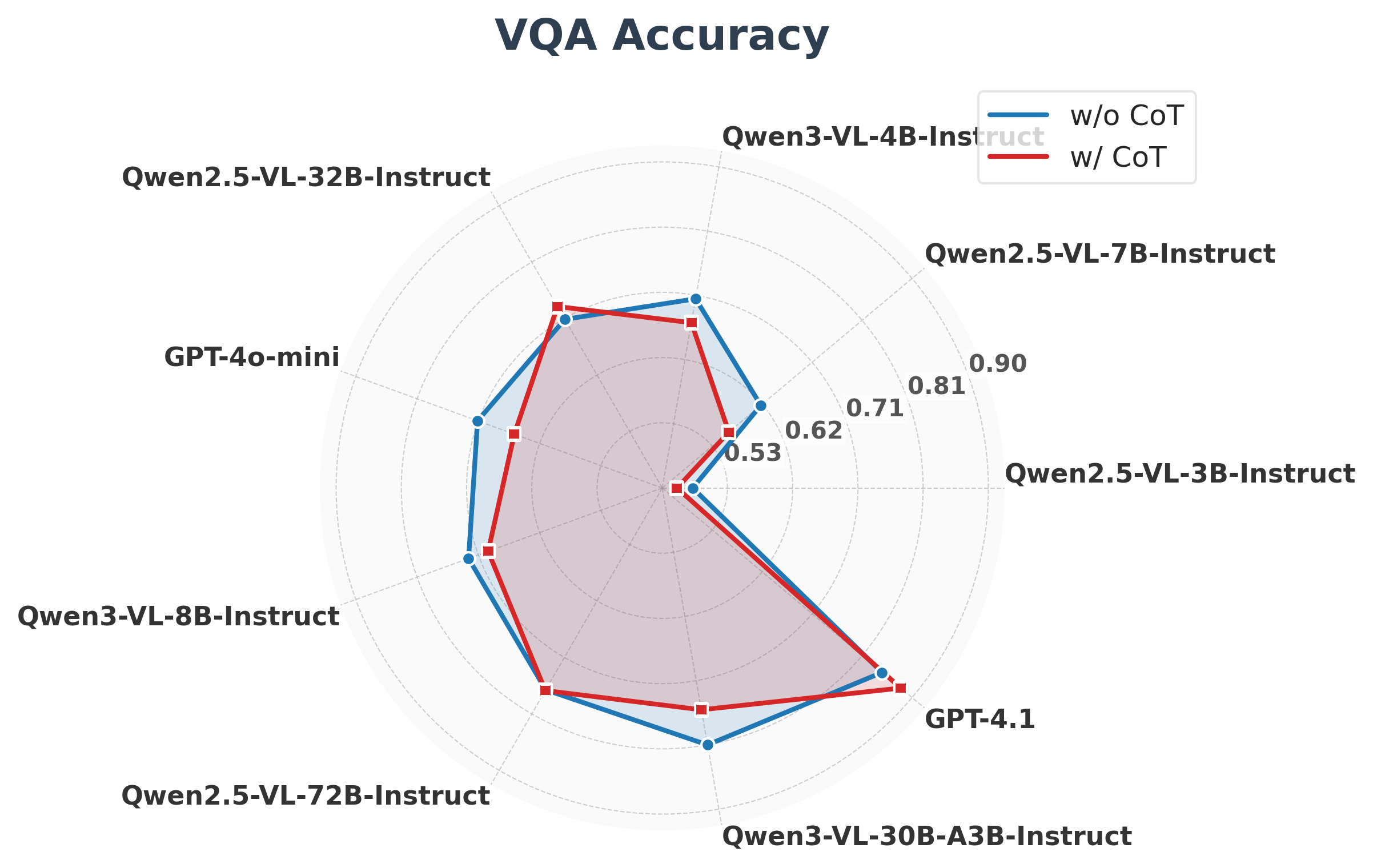}
\caption{Impact of Chain-of-Thought on VQA. A radar chart comparing the VQA Accuracy of various models with and without Chain-of-Thought (CoT) prompting, showing mixed effectiveness depending on the model scale and capability.}
\label{fig:vqa_accuraccy}
\end{figure}

We visualize the impact of CoT prompting on model performance. Figure~\ref{fig:classification_acc} and Figure~\ref{fig:vqa_accuraccy} illustrate the accuracy shifts in Classification and VQA tasks, respectively, when CoT is enabled versus disabled

\subsection{Performance on the English-Translated Dataset}

\begin{table*}[tbp!]
  \centering
  \scriptsize
  \setlength{\tabcolsep}{4pt}
  \renewcommand{\arraystretch}{1.1} 
  \begin{adjustbox}{width=0.95\textwidth,center}
    \begin{tabularx}{\textwidth}{
      l
      >{\raggedleft\arraybackslash}X  % Classification
      >{\raggedleft\arraybackslash}X  % Nutrition MAE
      >{\raggedleft\arraybackslash}X  % Nutrition RMSE
      >{\raggedleft\arraybackslash}X  % Nutrition MAPE
      >{\raggedleft\arraybackslash}X  % VQA
    }
    \toprule
    \multirow{2}{*}{\textbf{Model}} 
      & \multicolumn{1}{c}{\cellcolor{cyan!10}\textbf{Classification}} 
      & \multicolumn{3}{c}{\cellcolor{orange!10}\textbf{Nutrition Estimation}} 
      & \multicolumn{1}{c}{\cellcolor{purple!10}\textbf{VQA}} \\
    \cmidrule(lr){2-2} \cmidrule(lr){3-5} \cmidrule(lr){6-6}
      
      & \multicolumn{1}{c}{\tiny \textbf{ACC} $\uparrow$}
      & \multicolumn{1}{c}{\tiny Avg MAE $\downarrow$} 
      & \multicolumn{1}{c}{\tiny Avg RMSE $\downarrow$} 
      & \multicolumn{1}{c}{\tiny Avg MAPE $\downarrow$} 
      & \multicolumn{1}{c}{\tiny \textbf{ACC} $\uparrow$} \\
    \midrule

    % ================= Proprietary Models =================
    \rowcolor{gray!10}\multicolumn{6}{l}{\emph{\textbf{Proprietary Models}}} \\
    Gemini-2.5-Flash & 0.6779 & 71.98 & 114.92 & 41.09 & 0.8134 \\
    Gemini-2.5-Pro & 0.6959 & \underline{63.96} & \textbf{112.99} & 34.20 & \textbf{0.8706} \\
    Gemini-3-Flash-Preview & \textbf{0.7767} & \textbf{62.33} & \underline{113.87} & \textbf{27.83} & 0.8607 \\
    Gemini-3-Pro-Preview & \underline{0.7691} & 68.75 & 126.20 & \underline{28.13} & \underline{0.8644} \\
    GPT-4.1 & 0.6619 & 92.86 & 141.38 & 38.48 & 0.8371 \\
    GPT-4o & 0.6238 & 98.67 & 148.03 & 40.62 & 0.7736 \\
    GPT-4o-mini & 0.5170 & 97.04 & 147.62 & 41.93 & 0.7264 \\

    \midrule
    % ================= Open-source Models =================
    \rowcolor{gray!10}\multicolumn{6}{l}{\emph{\textbf{Open-Source Models}}} \\
    Qwen-2.5-VL-32B-Instruct & 0.5565 & 95.35 & 141.02 & 41.23 & 0.7077 \\
    Qwen-2.5-VL-3B-Instruct & 0.5094 & 110.43 & 156.67 & 51.21 & 0.4067 \\
    Qwen-2.5-VL-72B-Instruct & 0.6221 & 94.20 & 142.53 & 40.35 & 0.7251 \\
    Qwen-2.5-VL-7B-Instruct & 0.5881 & 108.42 & 154.71 & 46.15 & 0.6157 \\
    Qwen-3-VL-30B-A3B-Instruct & 0.6155 & 87.22 & 138.71 & 37.33 & 0.7687 \\
    Qwen-3-VL-30B-A3B-Thinking & 0.5843 & 88.63 & 138.78 & 42.04 & 0.7201 \\
    Qwen-3-VL-4B-Instruct & 0.5749 & 106.73 & 152.72 & 42.82 & 0.6766 \\
    Qwen-3-VL-4B-Thinking & 0.5426 & 111.66 & 158.05 & 51.19 & 0.6841 \\
    Qwen-3-VL-8B-Instruct & 0.5856 & 72.82 & 116.64 & 38.81 & 0.6741 \\
    Qwen-3-VL-8B-Thinking & 0.5340 & 114.16 & 160.55 & 50.05 & 0.6866 \\
    \bottomrule
    \end{tabularx}
  \end{adjustbox}
  \vspace{-0.5em}
  \caption{Performance on the English-Translated Dataset. Evaluation results for selected models on the English version of DiningBench, comparing Classification Accuracy, Nutrition Estimation metrics, and VQA Accuracy. The \textbf{best} and \underline{second-best} results are highlighted in bold and underlined, respectively.}
  \label{tab:english_main_results}
\end{table*}

To facilitate global adoption, we constructed a high-quality English version of DiningBench via Gemini-3-Pro-Preview translation and manual verification. Table~\ref{tab:english_main_results} presents the evaluation results.

\textbf{Classification declines due to semantic misalignment.} Fine-Grained Classification accuracy dropped universally across all models compared to the Chinese dataset. The Qwen series exhibited the most significant degradation (e.g., Qwen-3-VL-8B-Instruct dropped from 64.15\% to 58.56\%). This suggests a "semantic gap": models possess stronger multi-modal alignment for indigenous dish names encountered during pre-training, whereas translated English names may lack the specific cultural or visual associations required for fine-grained discrimination.

\textbf{Nutrition Estimation benefits from English prompts.} Conversely, Nutrition Estimation performance improved for a significant subset of models, including the Gemini-2.5 and GPT-4o series. This indicates that these models may possess more robust quantitative reasoning pathways or better grounding when processing English prompts, potentially due to the dominance of English in their pre-training corpora for reasoning tasks.

\subsection{Quality Assurance of DiningBench}
As detailed in Section~\ref{sec:dataset_construction}, DiningBench integrates a rigorous quality assurance pipeline. To empirically validate the dataset's reliability, we conducted an independent external audit.

We engaged three Ph.D. students from Humanities, Social Sciences, and STEM fields, independent of the dataset creation process. Using stratified random sampling, we selected 210 entries (70 per task). Evaluators applied the strict inclusion criteria used during construction to judge each sample. The audit resulted in a 100\% approval rate across all evaluators, confirming the exceptional quality and validity of the DiningBench dataset.

\begin{table*}[htbp]
    \centering
    \begin{tabular}{p{15cm}}
    \toprule
    Please strictly follow the \textbf{Chain of Thought} steps below for the analysis: \\
    \\
    1. \textbf{Visual Observation}: Describe the food's color, shape, texture, plating characteristics, and inferred cooking state. \\
    \\
    2. \textbf{Ingredient Breakdown}: Identify and list the main ingredients, auxiliary ingredients, seasonings, and sauce components in detail. \\
    \\
    3. \textbf{Comprehensive Answer}: Answer the question based on the original image content and the observations and breakdown above. \\
    \bottomrule
    \end{tabular}
    \caption{Prompt Template for Chain-of-Thought (CoT). The structured prompt used to guide models in generating step-by-step reasoning, covering visual observation, ingredient breakdown, and comprehensive answering.}
    \label{tab:food_analysis_prompt}
\end{table*}

\section{Prompts}
\label{app:prompts}

In this Section, we provide the prompts we used.

\begin{table*}[htbp]
    \centering
    \begin{tabular}{p{15cm}}
    \toprule
    You are an expert in nutrition and health science. Your task is to estimate the calories, macronutrients, health metrics, and trace element content of food based on the provided information. \\
    \\
    In addition to the text information in \texttt{food\_info}, you will be provided with the merchant's promotional image (\texttt{food\_picture}) and a user-uploaded image (\texttt{user\_picture}). These may contain dish details, ingredients, cooking methods, or specific calorie counts. You must synthesize all available information. \\
    \\
    Pay close attention to the following fields and logic: \\
    1. \texttt{food\_unit}: Indicates the measurement unit. \\
    2. \texttt{food\_description}: The merchant's description. It may contain reference calorie info, but you must adhere to the following rules to obtain real, objective, and precise values: \\
    \quad - Determine if the stated calories are for the whole dish or per 100g. If per 100g, calculate the total based on \texttt{food\_unit}. \\
    \quad - If there are different specifications/sizes, use the data corresponding to the selected specification. \\
    \quad - If Protein, Carbs, or Fat data is missing, you must generate these necessary values using your professional knowledge. \\
    \quad - \textbf{CRITICAL THINKING}: Merchants often under-report calories/fat or over-report protein for marketing purposes. You must detect such false advertising or underestimation. If detected, completely discard the \texttt{food\_description} numerical data and generate values based on your world knowledge and the \texttt{food\_info} ingredients. \\
    \\
    For \texttt{food\_picture} and \texttt{user\_picture}: \\
    \quad 1) Do not overlook decimal points (.) in numerical values. \\
    \quad 2) When uncertain, cross-reference with \texttt{food\_info} and description. \\
    \quad 3) If calorie info exists in both picture and description, choose the source that is more comprehensive. \\
    \\
    Food Metadata: \texttt{food\_info}: \textless\%s\textgreater. \\
    \\
    Output strictly according to the following JSON format defined in \texttt{output\_format}. Do not output any other characters (no markdown, no explanations), ensure I can directly use \texttt{json.loads()}! \\
    \\
    \texttt{output\_format = \{} \\
    \texttt{  "refined\_dish\_name": "Refine the original dish name from food\_info to be concise without changing its meaning",} \\
    \texttt{  "calories": 0,} \\
    \texttt{  "protein": 0,} \\
    \texttt{  "carbohydrates": 0,} \\
    \texttt{  "fat": 0,} \\
    \texttt{  "calorie\_source": "Select one from: ['total\_in\_info', 'per\_100g\_in\_info', 'inferred'] based on the actual situation",} \\
    \texttt{  "micronutrients": "Select the top 2 most abundant elements from this list and concatenate them: [Vitamin C, Vitamin A, Vitamin B, Vitamin E, Calcium, Magnesium, Zinc, Dietary Fiber]"} \\
    \texttt{\}} \\
    \bottomrule
    \end{tabular}
    \caption{Prompt for Nutrition Estimation. The detailed instruction set provided to models for estimating nutritional content, including rules for handling metadata, unit conversion, and visual inference.}
    \label{tab:nutrition_prompt}
\end{table*}

\begin{table*}[htbp]
    \centering
    \begin{tabular}{p{15cm}}
    \toprule
    \textbf{\# Role} \\
    You are a ``Food Knowledge Graph Expert'' proficient in standard naming conventions for Chinese and Western cuisine, as well as a ``Visual Dataset Construction Expert.'' \\
    \\
    \textbf{\# Goal} \\
    Your task is to perform name standardization on the [Target Food Item] and [Initial Distractor Candidates], and ultimately output exactly \textbf{7} of the most visually deceptive distractors. \\
    \\
    \textbf{\# Inputs} \\
    - Raw Target Food Item: \{food\_info\} \\
    - List of Initial Distractor Candidates: \{spu\_items\_str\} \\
    \\
    \textbf{\# Workflow (Chain of Thought)} \\
    1. \textbf{Normalization}: \\
    \ \ - Convert ``Merchant Marketing Names'' into ``Universal Standard Dish Names''. \\
    \ \ - Rules: Remove modifiers (e.g., ``Secret'', ``Grandma's'', ``Signature'') and portion descriptions. \\
    \ \ - Keep: Core ingredients, cooking methods, and necessary cutting forms. \\
    2. \textbf{Filter \& Deduplicate}: \\
    \ \ - Compare the standardized [Initial Distractor Candidates] with the standardized [Target Food Item]. \\
    \ \ - Exclude items that have the exact same name as the target. \\
    \ \ - Exclude items that have obviously huge visual differences. \\
    \\
    \textbf{\# Constraints} \\
    1. \textbf{Quantity Constraint}: The \texttt{distractors} list must strictly contain \textbf{7} items. \\
    2. \textbf{Format Constraint}: Output only a raw JSON string. Do not include markdown markers. \\
    \\
    \textbf{\# Output Format} \\
    Output a standard JSON object: \\
    \texttt{\{} \\
    \texttt{  "target\_standard\_name": "Standardized Name of Target Food",} \\
    \texttt{  "distractors": [} \\
    \texttt{    \{} \\
    \texttt{      "standard\_name": "Distractor Standard Name 1",} \\
    \texttt{      "is\_created": false,} \\
    \texttt{      "reason": "Briefly describe the reason for selection..."} \\
    \texttt{    \},} \\
    \texttt{    ... (Total of 7 items)} \\
    \texttt{  ]} \\
    \texttt{\}} \\
    \bottomrule
    \end{tabular}
    \caption{Prompt for Distractor Generation. The prompt used to standardize food names and generate ``hard'' negative candidates from the same menu for the Fine-Grained Classification task.}
    \label{tab:distractor_prompt}
\end{table*}

\begin{table*}[htbp]
    \centering
    \begin{tabular}{p{15cm}}
    \toprule
    \textbf{\# Role Definition} \\
    You are a world-class \textbf{Food Science and Computer Vision Benchmark Architect}. Your expertise lies in designing highly challenging Multimodal VQA tasks to evaluate AI models on fine-grained visual recognition, nutritional reasoning, and logical inference. \\
    \\
    \textbf{\# Goal} \\
    Based on the provided \textbf{4 reference images} (Visual Input) and \textbf{[God-view Metadata]}, construct \textbf{1} high-difficulty VQA test sample. \\
    \\
    \textbf{\# Input Data} \\
    1. \textbf{Visual Data}: 4 uploaded food images (labeled \texttt{image\_1} (Merchant Promotional Image), \texttt{image\_2} (User Real Shot), \texttt{image\_3}, \texttt{image\_4}). \\
    2. \textbf{God-view Metadata}: \{food\_info\} \\
    \\
    \textbf{\# Task Configuration} \\
    \textbf{[Standard vs. Reality Discrepancy Audit]}: Select \texttt{image\_1} and at least one real shot. Compare the ``promotional promise'' vs. ``actual delivery'' regarding portion size, integrity, or freshness, and infer the specific consequences of this deviation (e.g., calorie gap or experience degradation). \\
    \\
    \textbf{\# Constraints \& Rules (Critical - MUST FOLLOW)} \\
    1. \textbf{Atomic Query Principle}: \\
    \ \ - \textbf{Strictly Forbidden}: Including multiple sub-questions (e.g., ``What is this? How many calories? Is it healthy?''). \\
    \ \ - The question must be \textbf{single-focused}, targeting only \textbf{one} core difficulty. Ensure it ends with a single question mark. \\
    2. \textbf{Visual-Agnostic Phrasing}: \\
    \ \ - \textbf{Strictly Forbidden}: Describing image content in the question. Do not say ``Based on the golden crispy crust...''; instead ask ``Evaluate the texture characteristics of the food surface...''. \\
    \ \ - The question must force the model to \textbf{``look''} for itself. If the question reveals colors, shapes, or ingredient names, the test is a failure. \\
    \ \ - Do not leak specific values from the Metadata. \\
    3. \textbf{Inference Depth}: \\
    \ \ - Reject simple ``captioning'' style recognition. The question must involve implicit reasoning logic. \\
    4. \textbf{Fact-Grounded Answer}: \\
    \ \ - While the question cannot contain Metadata, the \textbf{Answer} and \textbf{CoT} must align with the [God-view Metadata] to ensure numerical precision and prevent hallucinations. \\
    \\
    \textbf{\# Output Format} \\
    Output only a standard JSON object. Do not include markdown markers or explanatory text. \\
    \texttt{\{} \\
    \texttt{  "difficulty\_level": "Hard/Medium/Easy",} \\
    \texttt{  "image": ["image\_x", "image\_y"], // Precisely select images relevant to the issue} \\
    \texttt{  "question": "(Concise, no visual description, no metadata values, single question mark)",} \\
    \texttt{  "cot\_gt": "Chain-of-thought process required to answer...",} \\
    \texttt{  "final\_answer": "Concise conclusion (including key values or judgments)"} \\
    \texttt{\}} \\
    \bottomrule
    \end{tabular}
    \caption{Prompt for VQA Generation (Multi Image Analysis). The instruction used to generate high-difficulty VQA samples that focus on identifying discrepancies between promotional reference images and real user-uploaded photos.}
    \label{tab:vqa_prompt}
\end{table*}

\begin{table*}[htbp]
    \centering
    \begin{tabular}{p{15cm}}
    \toprule
    \textbf{\# Role Definition} \\
    You are a \textbf{``Senior Food Critic''} with Michelin-star level appreciation capabilities. Your core competency lies in combining fine-grained \textbf{Visual Cues} with a deep \textbf{Culinary Knowledge Graph} to perform logically rigorous provenance reasoning. \\
    \\
    \textbf{\# Task Objective} \\
    Based on the provided \textbf{Food Image Metadata}, construct \textbf{5-8 High-Difficulty} VQA data samples. \\
    \textbf{Note: The output must be in strict JSON Array format.} \\
    \\
    \textbf{\# Difficulty Criteria (What is ``High-Difficulty''?)} \\
    Reject simple object recognition questions (e.g., ``What dish is this?''). You must adhere to the following standards: \\
    1. \textbf{Strong Visual Dependency}: The answer cannot be obtained solely by reading the metadata; it must describe visual features mentioned (e.g., Maillard reaction, glossiness, stacking order, oil state). \\
    2. \textbf{Multi-step Logic Chain}: Question $\rightarrow$ Observe specific features $\rightarrow$ Combine with culinary principles $\rightarrow$ Rule out distractors $\rightarrow$ Conclusion. \\
    3. \textbf{Detail Sensitivity}: Must distinguish between extremely similar states (e.g., Are the spring onions ``raw and crisp'' or ``wilted from hot oil''? Is the sauce ``drizzled'' on top or ``stewed'' in?). \\
    \\
    \textbf{\# Required Categories} \\
    1. \textbf{Cooking Technique \& State Reverse-Engineering}: Infer specific cooking methods (pan-fry, deep-fry, roast, steam, sous-vide) from surface textures (Maillard reaction, dehydration shrinkage, emulsification). \\
    2. \textbf{Dietary Restrictions \& Ingredient Audit}: Visual verification for specific groups (Keto, Vegan, Allergy). E.g., Judging if it is a vegan substitute based on cheese stretch or oil separation. \\
    3. \textbf{Image-Text Consistency \& Counterfactual Reasoning}: Verify if metadata conflicts with visual performance, or ask ``How would the taste change if [specific visual feature] were missing?''. \\
    \\
    \textbf{\# Output Format} \\
    Strictly output a valid \textbf{JSON List} format. Do not include Markdown markers or any introductory/concluding text. \\
    \\
    \textbf{Target Structure Example:} \\
    \texttt{\{} \\
    \texttt{  "vqa\_samples": [} \\
    \texttt{    \{} \\
    \texttt{      "category": "Cooking Technique \& State Reverse-Engineering",} \\
    \texttt{      "question": "Observing the color gradient and crust thickness...",} \\
    \texttt{      "answer": "It was not fully rested.",} \\
    \texttt{      "visual\_cues": ["Blood seepage at the bottom", "Uneven center color"],} \\
    \texttt{      "reasoning": "First, significant myoglobin leakage indicates..."} \\
    \texttt{    \},} \\
    \texttt{    ...} \\
    \texttt{  ]} \\
    \texttt{\}} \\
    \\
    \textbf{\# Data Input} \\
    \textbf{Metadata:} \\
    \{food\_info\} \\
    \bottomrule
    \end{tabular}
    \caption{Prompt for Culinary VQA Generation. The prompt designed for constructing complex reasoning questions related to cuisine technique, dietary suggestion, and counterfactual reasoning.}
    \label{tab:culinary_vqa_prompt}
\end{table*}

\begin{table*}[htbp]
    \centering
    \begin{tabular}{p{15cm}}
    \toprule
    \textbf{Role Definition} \\
    You are a senior \textbf{Fine-grained Visual Recognition Data QA Specialist}. You need to audit the validity of a ``Food Recognition'' multiple-choice question. \\
    This question aims to test the model's ability to distinguish visually similar dishes; therefore, distractors are allowed to be highly visually similar to the correct answer. \\
    \\
    \textbf{[Question Data]} \\
    1. Candidate Options: \\
    \{option\_str\} \\
    2. Standard Answer (Ground Truth): \\
    \{gt\_letter\}. \{correct\_name\} \\
    \\
    \textbf{[Pass Criteria - Must meet all]} \\
    1. \textbf{Image Quality}: The image is clear with a distinct subject. \\
    2. \textbf{Correct GT}: The standard answer must correctly describe the food in the image. \\
    3. \textbf{Unique Answer}: Although distractors may look very similar (e.g., ``Braised Beef Noodles'' vs. ``Spicy Beef Noodles''), they must be \textbf{incorrect} descriptions. If a distractor serves as a valid label for the image (i.e., multiple correct answers exist), it is invalid. \\
    4. \textbf{Difficulty}: If the gap between the standard answer and other categories is too large (making the question too simple), it does not pass. \\
    \\
    \textbf{[Failure Examples]} \\
    - Error Case A (Label Error): Image is ``Burger'', but GT is ``Sandwich''. \\
    - Error Case B (Multi-Solution): Image is ``Stir-fried Potato Strips'', Option A is ``Stir-fried Potato Strips'' (GT), Option B is ``Fried Potato Strips'' (also correct). \\
    \\
    \textbf{[Output Format]} \\
    Please output only a valid JSON object, do not output markdown markers: \\
    \texttt{\{} \\
    \texttt{  "is\_valid": true, // true only if image is clear, GT is correct, and answer is unique} \\
    \texttt{  "analysis": "Brief analysis. If high-difficulty distractors exist, note 'Distractors are confusing but GT is unique, question valid'.",} \\
    \texttt{  "error\_type": "None" // Options: "Wrong\_GT", "Multi\_Correct", "Bad\_Image", "Too\_Easy"} \\
    \texttt{\}} \\
    \bottomrule
    \end{tabular}
    \caption{Quality Assurance Prompt for Classification. The criteria and instructions used by AI auditors to validate the quality, difficulty, and uniqueness of Fine-Grained Classification samples.}
    \label{tab:data_qa_prompt}
\end{table*}

\begin{table*}[htbp]
    \centering
    \begin{tabular}{p{15cm}}
    \toprule
    \textbf{Role Definition} \\
    You are a \textbf{Food Nutrition Data Final Audit Expert} with ``Absolute Zero Tolerance'' standards. Your task is to clean the ``Food Nutrition Estimation'' Golden Test Set. \\
    Any data with flaws, ambiguity, logical conflicts, or estimation bias must be ruthlessly rejected. Your goal is to ensure the remaining data is 100\% perfect and accurate. \\
    \\
    \textbf{[Data to Audit]} \\
    Standard Answer (Ground Truth): \\
    \{cal\_response2\_str\} \\
    (Also reference the input food image) \\
    \\
    \textbf{[Audit Process \& Absolute Rejection Criteria]} \\
    Please execute the following 3 checks in order. If any rejection criterion is met, immediately flag as invalid (false): \\
    \\
    1. \textbf{Image Validation} \\
    \ \ - \textbf{Reject if}: Image is blurry, over/under-exposed, subject is largely occluded, contains irrelevant clutter, or is not real food (e.g., painting, model). \\
    \\
    2. \textbf{Mathematical Logic Audit} \\
    \ \ - \textbf{Core Formula}: Verify using the Atwater system: $ E \approx 4 \times P + 4 \times C + 9 \times F $ \\
    \ \ \ \ (Where $E$=Energy/kcal, $P$=Protein/g, $C$=Carbs/g, $F$=Fat/g). \\
    \ \ - \textbf{Reject if}: If macronutrients are provided in the GT, calculate the theoretical calories. If the error between labeled Total Calories and theoretical calculation exceeds \textbf{10\%}, it is a logical error. \\
    \ \ - \textbf{Reject if}: Values violate physical common sense (e.g., Sum of Carbs + Protein + Fat in 100g food exceeds 100g). \\
    \\
    3. \textbf{Visual-Data Matching} \\
    \ \ - \textbf{Reject if}: \textbf{Nutritional Density Mismatch}. The labeled values fail to reflect the physical form of the food in the image (e.g., image shows fatty pork belly but labeled fat is extremely low; or image is pure vegetable salad but labeled calories are 800kcal). \\
    \\
    \textbf{[Output Requirements]} \\
    1. Output only a pure JSON object. Strictly NO markdown code blocks. \\
    2. Even if a minor flaw is found, judge as false. We need perfect data. \\
    \\
    \textbf{[JSON Field Definitions]} \\
    - \texttt{is\_valid} (bool): \texttt{true} only implies the data is flawless; otherwise \texttt{false}. \\
    - \texttt{error\_type} (str): Select the primary error type: \\
    \ \ \ \ - \texttt{"None"}: Data is perfect. \\
    \ \ \ \ - \texttt{"Bad\_Image"}: Image quality issue. \\
    \ \ \ \ - \texttt{"Data\_Logic\_Error"}: Nutrients and Total Calories are mathematically inconsistent. \\
    \ \ \ \ - \texttt{"Visual\_Value\_Conflict"}: Visual portion/features conflict seriously with values. \\
    - \texttt{analysis} (str): Briefly explain the rejection reason. If it is a numerical issue, list specific comparison data (e.g., ``Calculated calories 300kcal, labeled 150kcal, error too large''). \\
    \bottomrule
    \end{tabular}
    \caption{Quality Assurance Prompt for Nutrition Data. The rigorous audit protocol for validating nutrition samples, ensuring image quality, mathematical consistency (Atwater system), and visual-data matching.}
    \label{tab:nutrition_qa_prompt}
\end{table*}

\begin{table*}[htbp]
    \centering
    \begin{tabular}{p{15cm}}
    \toprule
    \textbf{\# Role} \\
    You are a \textbf{Chief VQA Auditor}. Your task is to build a \textbf{``High-Difficulty''} test set for SOTA models. \\
    Your core principles are: \textbf{Atomic, High-Difficulty, Zero-Tolerance}. Any mediocre, simple, structurally chaotic, or verbose data must be immediately ``sentenced to death''. \\
    \\
    \textbf{\# Input Data} \\
    - Visual Context: [Model analysis based on image] \\
    - Question: \{question\} \\
    - Answer: \{answer\} \\
    - Reasoning: \{reasoning\} \\
    \\
    \textbf{\# Critical Audit Standards (Strict Veto System)} \\
    Please execute the following audits in order. If any violation is found,  rule \texttt{is\_valid: false}: \\
    \\
    1. \textbf{Structural Integrity}: \\
    \ \ - \textbf{No Multi-Part Questions}: A question must be \textbf{atomic}, asking only one core point. Any question containing conjunctions (e.g., ``Where is... \textbf{AND} what is he doing?'', ``What is the color \textbf{and} shape?'') must be \textbf{rejected}. \\
    \ \ - \textbf{False Premise}: Questions containing information not present in the image (e.g., asking about a red car when the car is blue) must be \textbf{rejected}. \\
    \\
    2. \textbf{Fact \& Vision Zero-Tolerance}: \\
    \ \ - \textbf{Visual Hallucination}: Objects must be \textbf{clearly visible}. If occluded, truncated, blurry, or requiring excessive guessing, \textbf{reject}. \\
    \ \ - \textbf{Unanswerable}: Questions where no definite conclusion can be drawn based on current visual info. \\
    \\
    3. \textbf{Difficulty \& IQ Filter (The ``Triviality'' Trap)}: \\
    \ \ - \textbf{Reject Low-Level Tasks}: If a non-expert human can answer in \textbf{$<$0.5s} without thinking (e.g., simple color recognition, counting large objects, obvious OCR), it is trash data. \textbf{Reject}. \\
    \ \ - \textbf{Keep Standard}: Only retain samples requiring multi-step reasoning, fine-grained distinction (e.g., specific dog breed vs just ``dog''), or commonsense reasoning. \\
    \\
    4. \textbf{Answer Quality Control}: \\
    \ \ - \textbf{Reject Verbosity}: Answers must be \textbf{minimalist}. No redundancy. \\
    \ \ - \textbf{Reject Subjectivity}: Answers must be objective and unique. If different people would give different answers, \textbf{reject}. \\
    \\
    \textbf{\# Output Schema} \\
    Output a strict JSON object. Do not use Markdown markers. \\
    \texttt{\{} \\
    \texttt{  "analysis\_trace": "Concise analysis process...",} \\
    \texttt{  "is\_valid": boolean, // Final verdict. False if any criteria met.} \\
    \texttt{  "quality\_score": integer, // Strict score (0-10). <6 is trash.} \\
    \texttt{  "issue\_category": string, // Enums: "Multi\_Part\_Question", "Trivial\_Task", "False\_Premise", "Unanswerable\_Visual", "Verbose\_Answer", "Subjective\_Ambiguous", "None"} \\
    \texttt{  "correction\_suggestion": string // Mostly "Discard". Only suggest fix if question is perfect but answer has minor formatting issues.} \\
    \texttt{\}} \\
    \bottomrule
    \end{tabular}
    \caption{Quality Assurance Prompt for VQA. The strict audit guidelines used to filter VQA samples, rejecting trivial, multi-part, or hallucinated questions to ensure high benchmark difficulty.}
    \label{tab:vqa_audit_prompt}
\end{table*}

\end{document}